\documentclass[10pt,twocolumn,letterpaper]{article}

\usepackage{cvpr}              

\usepackage{graphicx}
\usepackage{amsmath}
\usepackage{amssymb}
\usepackage{booktabs,eucal}

\usepackage{ upgreek } 

\usepackage{multirow} 

\usepackage[dvipsnames,table,xcdraw]{xcolor}
\usepackage[pagebackref=false,breaklinks,colorlinks,citecolor=RoyalBlue,bookmarks=false]{hyperref}

\usepackage[capitalize]{cleveref}
\crefname{section}{Sec.}{Secs.}
\Crefname{section}{Section}{Sections}
\Crefname{table}{Table}{Tables}
\crefname{table}{Table}{Tables}

\def\cvprPaperID{6109} 
\def\confName{CVPR}
\def\confYear{2023}

\begin{document}

\title{
IT3D: Improved Text-to-3D Generation with Explicit View Synthesis
}

\author{\textbf{Yiwen Chen\thanks{Equal contribution.}$^{\ \,1,2}$} \quad \textbf{Chi Zhang\footnotemark[1]$^{\ \,3}$} \quad \textbf{Xiaofeng Yang$^2$} \quad \textbf{Zhongang Cai$^4$} \\ \quad \textbf{Gang Yu$^3$} \quad 
\textbf{Lei Yang$^4$} \quad \textbf{Guosheng Lin\thanks{Corresponding Authors.}$^{\ \,1,2}$}
\vspace{0.2cm} \\
$ ^1$S-Lab, Nanyang Technological University \\
$ ^2$School of Computer Science and Engineering, Nanyang Technological University \\
$ ^3$Tencent PCG, China; \quad
$ ^4$SenseTime Research \\
{\fontsize{10}{10}\selectfont \url{https://github.com/buaacyw/IT3D-text-to-3D}}
}

\maketitle


\begin{abstract}
Recent strides in Text-to-3D techniques have been propelled by distilling knowledge from powerful large text-to-image diffusion models (LDMs). Nonetheless, existing Text-to-3D approaches often grapple with challenges such as over-saturation, inadequate detailing, and unrealistic outputs. This study presents a novel strategy that leverages explicitly synthesized multi-view images to address these issues. Our approach involves the utilization of image-to-image pipelines, empowered by LDMs, to generate posed high-quality images based on the renderings of coarse 3D models. Although the generated images mostly alleviate the aforementioned issues, challenges such as view inconsistency and significant content variance persist due to the inherent generative nature of large diffusion models, posing extensive difficulties in leveraging these images effectively. To overcome this hurdle, we advocate integrating a discriminator alongside a novel Diffusion-GAN dual training strategy to guide the training of 3D models. For the incorporated discriminator, the synthesized multi-view images are considered real data, while the renderings of the optimized 3D models function as fake data. We conduct a comprehensive set of experiments that demonstrate the effectiveness of our method over baseline approaches.

\end{abstract}


\begin{figure*}[h!]
    \centering
    \includegraphics[width=\linewidth,trim={0cm 10cm 0.0cm 0.0cm}, clip]{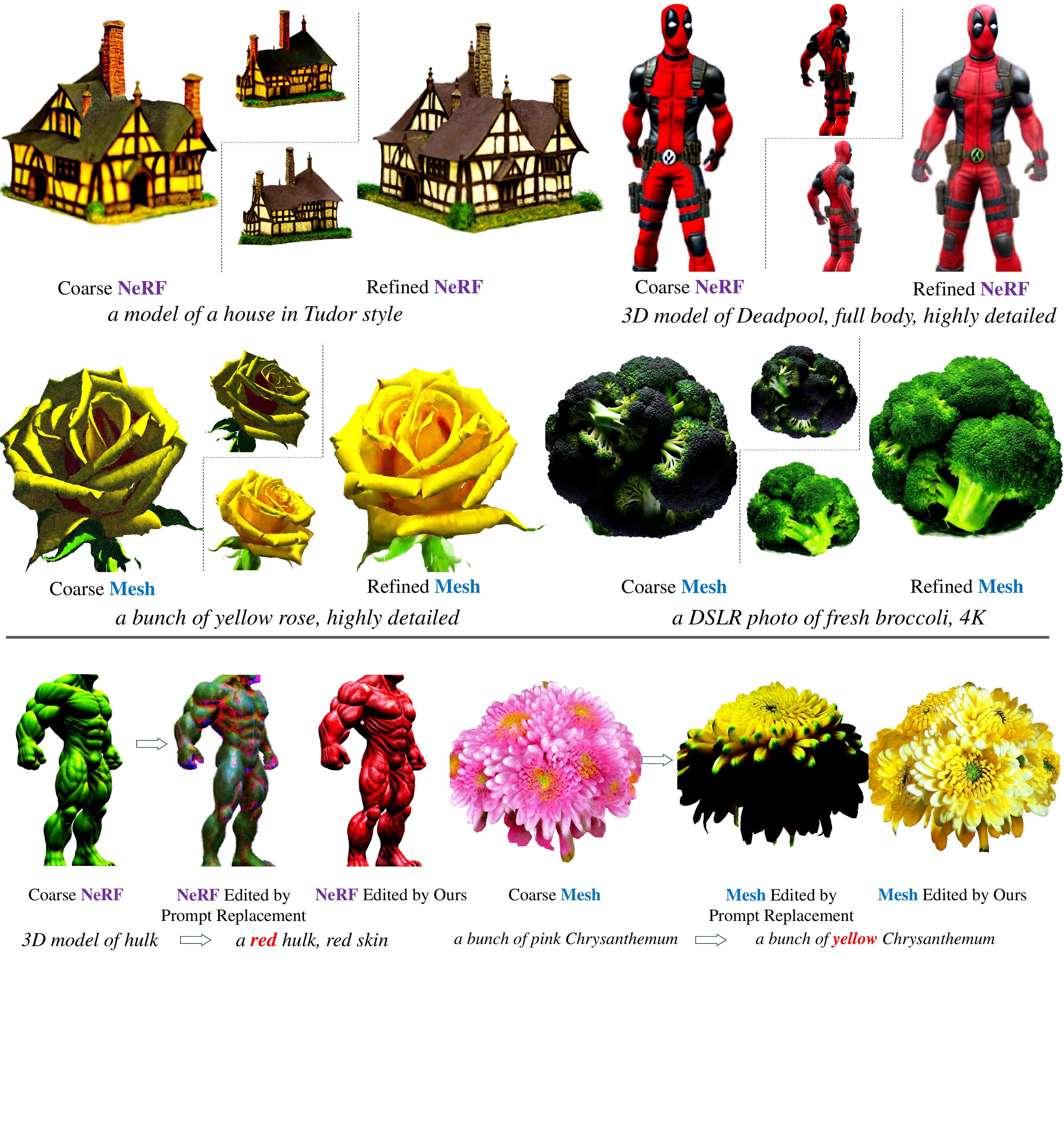}
    \caption{
    \textbf{Results and applications of IT3D.}
    \textbf{Top: high-resolution text-to-3D Refinement}.
    IT3D can refine the texture and geometry of 3D models generated by diffusion prior.
    \textbf{Bottom: high-resolution prompt-based editing}. Our method can edit 3D models given a prompt. Naively editing 3D models by prompt replacement would potentially cause an editing failure.}
    \label{fig:f1}
\end{figure*}

\section{Introduction}

In recent years, the field of text-to-image has witnessed remarkable progress, sparking a surge of interest within the research community to extend this advancement to the domain of 3D generation. 
This enthusiasm can largely be attributed to the emergence of methods capitalizing on pre-trained 2D text-to-image diffusion models~\cite{sohl2015deep,ho2020denoising,song2021scorebased}. 
By harnessing the immense volumes of training image data, these models acquire a comprehensive understanding of object appearance and geometry. 
Such advances have paved the way for text-to-3D generation without the explicit need for 3D datasets, showcasing an unparalleled combination of flexibility, diversity, and potential for cost and time savings.

A cornerstone in this domain is the innovative work of Dreamfusion~\cite{poole2022dreamfusion}. Their introduction of the Score Distillation Sampling (SDS) algorithm has been a game-changer, primarily due to its capability to generate diverse 3D objects using mere textual prompts~\cite{poole2022dreamfusion}. 
Despite its revolutionary approach, it comes with its set of challenges. A significant limitation is its control over the geometry and texture of the generated models, often leading to issues like over-saturation and the multi-face appearance of models (as depicted by coarse models in Fig.~\ref{fig:f1}). Furthermore, there is an observed inefficacy in improving the model quality by merely amplifying the textual prompts~\cite{poole2022dreamfusion,lin2022magic3d,wang2023prolificdreamer}.

To address these challenges, our research seeks to present an enhanced methodology for 3D generation. At its core, our method emphasizes the explicit synthesis of multi-view images of the intended 3D model, subsequently employing these images for 3D object reconstruction. 
This process begins by utilizing an existing text-to-3D generation model, such as DreamFusion~\cite{poole2022dreamfusion}, to craft a rudimentary depiction of the object. By rendering these foundational models, we gain an elemental representation of the object's geometry and spatial arrangement.
Building upon these initial renderings, our method refines the view images using an image-to-image (I2I) generation pipeline~\cite{zhang2023adding,brooks2023instructpix2pix,rombach2022high}.
By focusing on generating 2D images, a task that is simpler and more extensively studied, we can employ a range of techniques to enhance the image quality, resulting in the creation of more realistic and detailed images from different views of the 3D objects.

However, our approach is not without its unique set of challenges. The independent view generation  can lead to multi-view inconsistencies, such as differences in textures and geometries.
These inconsistencies pose a significant barrier to traditional 3D reconstruction methods and hinder them from producing satisfactory outcomes. 
To circumvent this obstacle, we introduce a discriminator paired with an adversarial loss to guide the learning of 3D models~\cite{goodfellow2014generative, chan2022efficient}. Here, we treat the enhanced multi-view images generated by I2I models as real data and the renderings of the optimized 3D models as fake data. By leveraging adversarial losses, our model effectively addresses these multi-view discrepancies, significantly enhancing the stability of training and improving the overall quality of the generated 3D objects.

Our method, termed IT3D, presents itself as a versatile plug-and-play solution, seamlessly integrating with any text-to-3D methodology that  draws upon 2D diffusion priors. 
Our method is versatile, supporting a range of 3D output representations such as meshes and NeRFs. An additional advantage is its capacity to efficiently modify the appearance of 3D models using text, with examples illustrated in Fig.~\ref{fig:f1}. Empirical evaluations further reveal that our approach can accelerate training convergence, which in turn, reduces the number of required training steps and results in comparable overall training time.

 We provide comprehensive experimental results, both qualitative and quantitative, to validate the effectiveness of our methods.
Our empirical findings  show that our proposed method significantly improves the baseline models in terms of texture detail, geometry, and fidelity between text prompts and the resulting 3D objects.

\section{Related Work}
In this section, we review the literature related to text-to-3D generation. Our examination centers on two principal areas: the emergence of diffusion models for image generation and the evolution of text-to-3D generation methods.

\subsection{Text-to-image diffusion models.}
The recent surge in generative model research has been prominently characterized by the rise of diffusion models. 
Incorporating textual prompts within the diffusion model has proven to be a critical step toward creating high-caliber text-to-image diffusion models. Renowned models in this domain such as GLIDE~\cite{nichol2021glide}, DALL·E 2~\cite{ramesh2021zero}, Imagen~\cite{saharia2022photorealistic}, and StableDiffusion~\cite{rombach2022high}, leverage textual prompts as guiding parameters during the image generation process. As a result, these models have demonstrated significant competence in creating high-quality images that mirror the descriptions provided by the accompanying text.

\subsection{Text-to-3D generation.}
The groundbreaking work of DreamFusion~\cite{poole2022dreamfusion} led the way by introducing the Score Distillation Sampling (SDS) method. This innovative technique optimizes 3D models using diffusion prior. 
Subsequent models built upon these pioneering developments, seeking to further enhance and refine the 3D generation process~\cite{lin2022magic3d,chen2023fantasia3d,wang2023prolificdreamer,zhu2023hifa,wang2022score,tang2023make,seo2023let}.
For instance, Magic3D~\cite{lin2022magic3d} sought to boost the resolution and learning speed of DreamFusion by utilizing a two-stage optimization strategy incorporating a sparse 3D hash grid structure. Similarly, DreamBooth3D integrated DreamBooth~\cite{ruiz2023dreambooth} and DreamFusion to facilitate the personalization of text-to-3D generative models using a few subject images. Additionally, ProlificDreamer~\cite{wang2023prolificdreamer} presented variational score distillation (VSD), a principled particle-based variational framework, to improve text-to-3D generation.

\begin{figure*}[h!]
    \centering
    \includegraphics[width=\linewidth,trim={0cm 2.5cm 0.0cm 0.0cm}, clip]{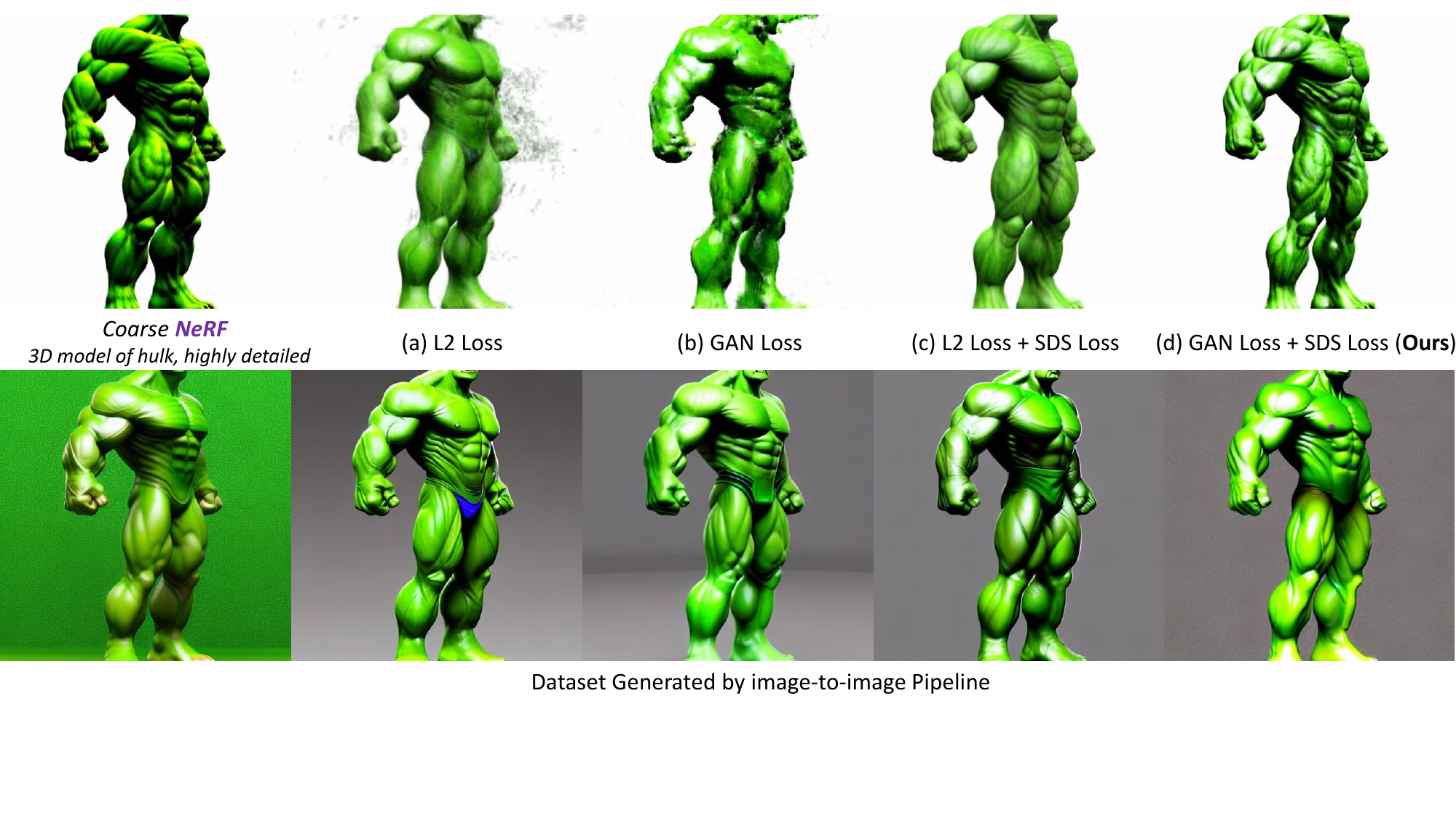}
    \caption{
    \textbf{Top: Results for different loss setting. Bottom: Data samples generated through image-to-image pipeline.}
    \textbf{Prompt: a 3D model of hulk, highly detailed, full body.}
    The generated dataset incorporate $60$ camera views and $5$ samples for each view. Given such a tiny but high variance dataset, (a) and (b) tend to result in noised refinement without the help of strong denoising ability of SDS loss. Comparing (c) and (d), (c) compromises to the average of generated dataset while (d) successfully bridge the distribution gap between coarse NeRF and generated dataset.
    }
    \label{fig:f2}
\end{figure*}

The work most closely related to ours is DreamBooth3D~\cite{raj2023dreambooth3d}. Similar to our approach, DreamBooth3D involves generating datasets based on renderings of coarse models and enhancing them using image-to-image pipelines. However, there are distinct differences between our methods and DreamBooth3D in two main aspects. Firstly, DreamBooth3D primarily focuses on image-to-3D and requires a few subject images to fine-tune a DreamBooth model~\cite{ruiz2023dreambooth}, resulting in a low-variance dataset for a single subject. In contrast, our method is geared towards enhancing text-to-3D generation. We generate high-variance datasets without the need for any tuning on large diffusion models. Secondly, our methods make more efficient use of the generated datasets by leveraging a 3D GAN. This approach proves superior to using an L2 loss with view and content-inconsistent datasets, as illustrated in Fig.~\ref{fig:f2}.

\begin{figure*}[!]
    \centering
    \includegraphics[width=\linewidth,trim={0cm 0.0cm 0.0cm 0.0cm}, clip]{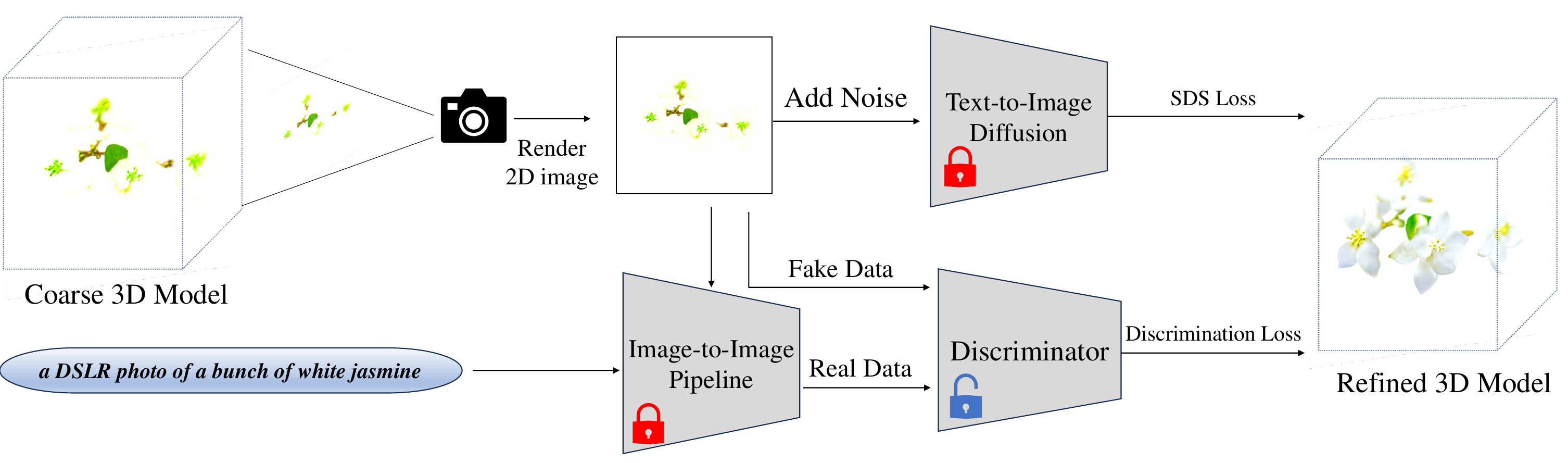}
    \caption{\textbf{Pipeline of IT3D.} Begin from a coarse 3D model, IT3D first generate a tiny posed dataset leveraging image-to-image pipeline conditioning on rendering of the coarse 3D model. Then incorporate a randomly initialized discriminator to distill knowledge form the generated dataset and update the 3D model with discrimination loss and SDS loss.}
    \label{fig:f0}
\end{figure*}
\section{Preliminary}
This section introduces the foundation upon which our approach is built. Specifically, we provide an overview of the DreamFusion model, a pivotal text-to-3D generation technique that forms our baseline. While we utilize DreamFusion~\cite{poole2022dreamfusion} predominantly in our experiments, it is essential to note that our design remains versatile and can be incorporated alongside other text-to-3D generation models.

The text-to-3D generation achieved by DreamFusion~\cite{poole2022dreamfusion} is composed of two key components: a neural scene representation, referred to as the 3D model, and a pre-trained text-to-image diffusion-based generative model that provides diffusion prior.

The 3D model, denoted as a parametric function $x = g(\theta)$, can produce images $x$ at specified camera poses. In this context, $g$ refers to the chosen volumetric renderer, while $\theta$ stands for a coordinate-based MLP representing a 3D volume. The diffusion model $\phi$ is accompanied by a learned denoising function $\epsilon_\phi(x_t;y,t)$, which predicts sampled noise $\epsilon$ based on the noisy image $x_t$, noise level $t$, and text embedding $y$. This denoising function supplies the gradient direction to update $\theta$, aiming to guide all rendered images towards high-probability density regions conditioned on the text embedding under the diffusion prior.

DreamFusion introduces Score Distillation Sampling (SDS) to compute the gradient as follows:
\begin{equation}
    \nabla_\theta \mathcal{L}_\text{SDS}(\phi, g(\theta)) = 
    \mathbb{E}_{t, \epsilon} \! \! \left[ w(t)(\epsilon_\phi(x_t;y,t) - \epsilon)\frac{\partial x}{\partial \theta} \right]. 
\end{equation}
Here, $w(t)$ denotes a weighting function. The scene model $g$ and diffusion model $\phi$ are modular elements within the framework, allowing for flexible choices. 

Despite the success of DreamFusion, over saturation, inadequate detailing, and unrealistic outputs can often be observed in text-to-3D method based on diffusion prior.

\section{Method}
This study aims to address prevalent issues in prior text-to-3D methods, namely over-saturation, lack of details, and unrealistic outputs~\cite{poole2022dreamfusion,lin2022magic3d,wang2023prolificdreamer}. While these challenges persist within the field of text-to-3D, substantial progress has been achieved in mitigating them by employing 2D image generation techniques driven by diffusion models. Building on this insight, we seek to tackle these concerns through the incorporation of 2D image generation techniques (Fig.~\ref{fig:f0}). We will show that high-quality 
multi-view images of 3D models that are free from the aforementioned issues can be generated at a low cost (Section~\ref{sec:data}).

However,  the intrinsic randomness of 2D generative models gives rise to content inconsistencies across different 
 views~\cite{zhang2023adding,rombach2022high}. 
This inherent nature leads to considerable difficulties in effectively leveraging these generated images. To harness the potential of the generated dataset more effectively, we propose an innovative approach involving the integration of a discriminator~\cite{chan2022efficient}. This discriminator serves to better distill knowledge from the generated datasets, thereby guiding the training process of the 3D model (Section~\ref{sec:gan}).

Despite the proficiency of 3D GANs in accommodating high variance datasets, the computational expense associated with training a 3D GAN to simulate realistic textures and intricate details renders it impractical for text-to-3D applications~\cite{chan2022efficient}. To expedite the training process, we confine the size of the generated dataset to approximately two hundred samples. However, this 
mini-dataset proves inadequate for training a discriminator 
to generate a 3D model with realistic textures and geometries
, as the discriminator quickly overfits to such a limited dataset (as depicted in Fig.~\ref{fig:f2}). To 
overcome this challenge, we introduce an innovative training strategy that employs a combination of diffusion prior and GAN loss (Section~\ref{sec:strategy}). This strategy results in a notable reduction in the training burden of the GAN component.

\subsection{Dataset Generation}
\label{sec:data}
In this section, we describe our data generation strategy in detail. IT3D starts with a coarse 3D model parameterized by $\theta$, derived using a text-to-3D baseline method that conditions on the prompt $T$. Firstly, we render the coarse 3D model across a range of camera poses represented by $C$, ultimately resulting in the creation of a dataset comprising posed images denoted as $D$. Subsequent to this, we employ image-to-image pipelines to generate a dataset of enhanced quality, termed $D'$, which is conditioned on both the prompt $T'$ and the original coarse dataset $D$. Users may either maintain $T'$ as equivalent to $T$ to refine the 3D model $\theta$, or select a distinct $T'$ to edit the 3D model $\theta$. It is noteworthy that IT3D does not necessitate a well-trained $\theta$. A preliminary 3D model exhibiting the foundational shape, albeit lacking extensive texture, is sufficient for providing depth maps or normal maps as conditioning images for the image-to-image pipeline. 

The collection of camera poses $C$ envelops the 3D model in a uniform manner, with the same number of images being generated from the image-to-image pipeline for each viewpoint. This uniform distribution of poses is crucial for achieving an unbiased pose distribution within the dataset, a factor of high importance for effective GAN training. In our experimentation, the size of $C$ ranges from 24 to 60, and 3 to 7 images are sampled for each viewpoint, contingent upon the complexity of the 3D model.

For the image-to-image pipeline, we opt for ControlNet~\cite{zhang2023adding} and Stable Diffusion~\cite{rombach2022high}. We extract feature maps from images in the coarse datasets, encompassing depth maps, normal maps, and soft-edge maps. These extracted features serve as conditioning images for ControlNet, ensuring the generation of data that retains viewpoint consistency. In the context of the Stable Diffusion image-to-image pipeline~\cite{rombach2022high}, random Gaussian noise is initially introduced to the rendered images. Subsequently, these noise-affected images are denoised with vanilla 2D diffusion denoising steps, ultimately yielding images of elevated quality.

\begin{figure*}[h!]
    \centering
    \includegraphics[width=\linewidth,trim={0cm 17.5cm 0.0cm 0.0cm}, clip]{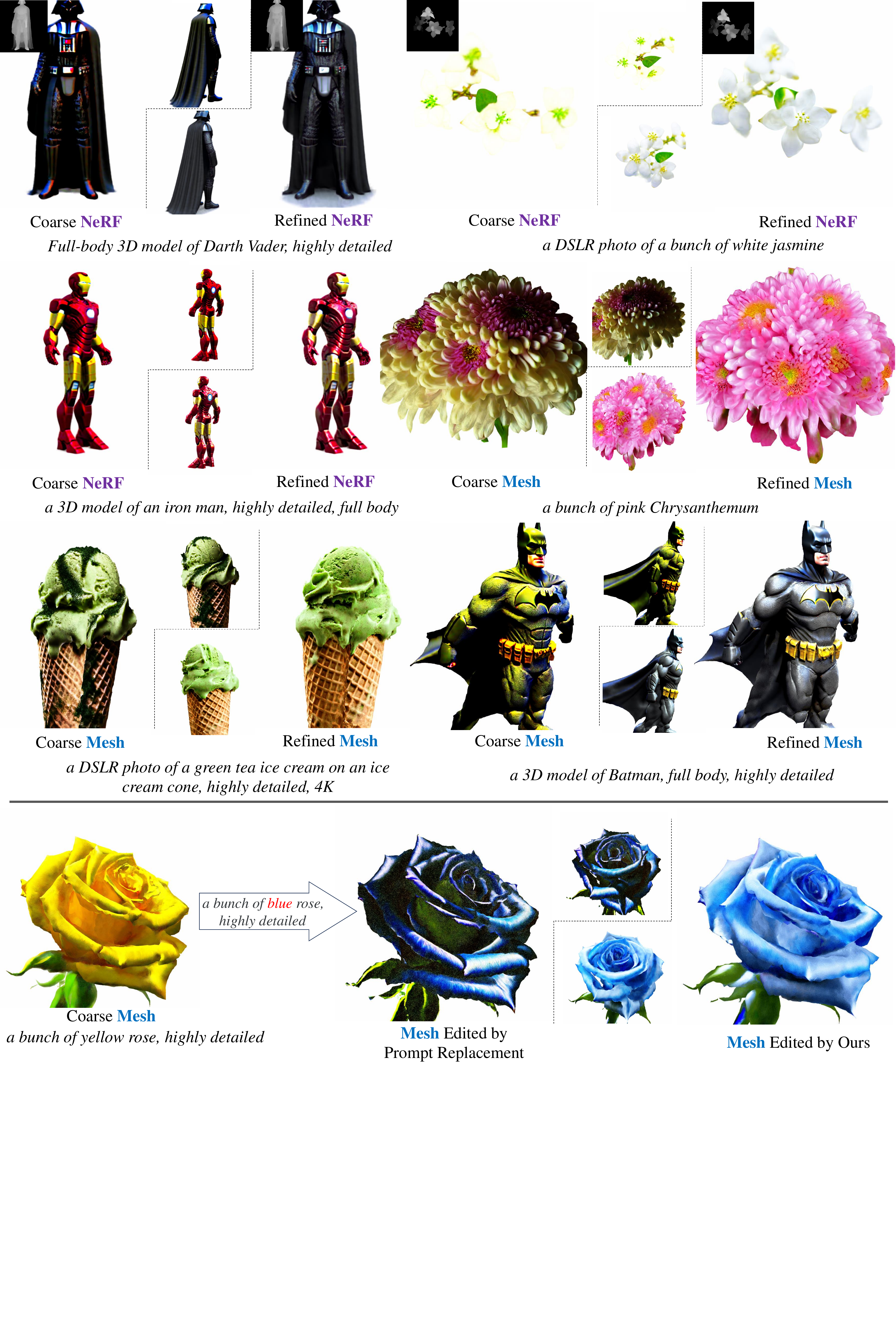}
    \caption{
    \textbf{More Results of IT3D.} \textbf{Top: high-resolution text-to-3D Refinement}.
    IT3D can refine the texture and geometry (body posture of Deadpool) 3D models generated by text-to-3D method. 
    \textbf{Bottom: high-resolution prompt-based editing.}
    IT3D can edit 3D models given a prompt. Naively editing NeRF by prompt replacement would potentially cause an editing failure.}
    \label{fig:f3}
\end{figure*}

\subsection{Handling Inconsistencies with GAN}
\label{sec:gan}

Although the image-to-image pipeline is conditioned on feature maps extracted from the original renderings during the generation process, it is inevitable that the generated dataset $D'$ exhibits inconsistencies in terms of content and camera views. Naively applying a plain L2 loss would yield suboptimal results due to its inclination to align the 3D model $\theta$ with the average content of the highly variable dataset (refer to Fig.~\ref{fig:f2}). 

However, the capabilities of Generative Adversarial Networks (GANs) shine in scenarios involving datasets characterized by high variance~\cite{chan2022efficient}. GANs have the ability to learn both geometry and texture-related knowledge from such datasets, subsequently guiding the 3D model to converge towards the same high-quality distribution exhibited by the generated dataset.

In our approach, we designate the 3D model as a 3D generator. We then integrate a discriminator initialized with random values. In this setup, the generated dataset $D'$ is treated as real data, while the renderings of the 3D model $\theta$ represent fake data. The role of the discriminator involves learning the distribution discrepancy between the renderings and $D'$, subsequently contributing to the discrimination loss which in turn updates the 3D model $\theta$.

\subsection{Diffusion-GAN Dual Training Strategy}
\label{sec:strategy}
Instead of relying solely on the discrimination loss, we propose an innovative Diffusion-GAN dual training strategy that combines the strengths of the diffusion prior and discrimination loss. This strategy utilizes the discrimination loss to guide the updating direction and leverages the diffusion prior to provide intricate geometry and texture details.
\begin{figure*}[h!]
    \centering
    \includegraphics[width=\linewidth,trim={0cm 0.5cm 0.0cm 0.0cm}, clip]{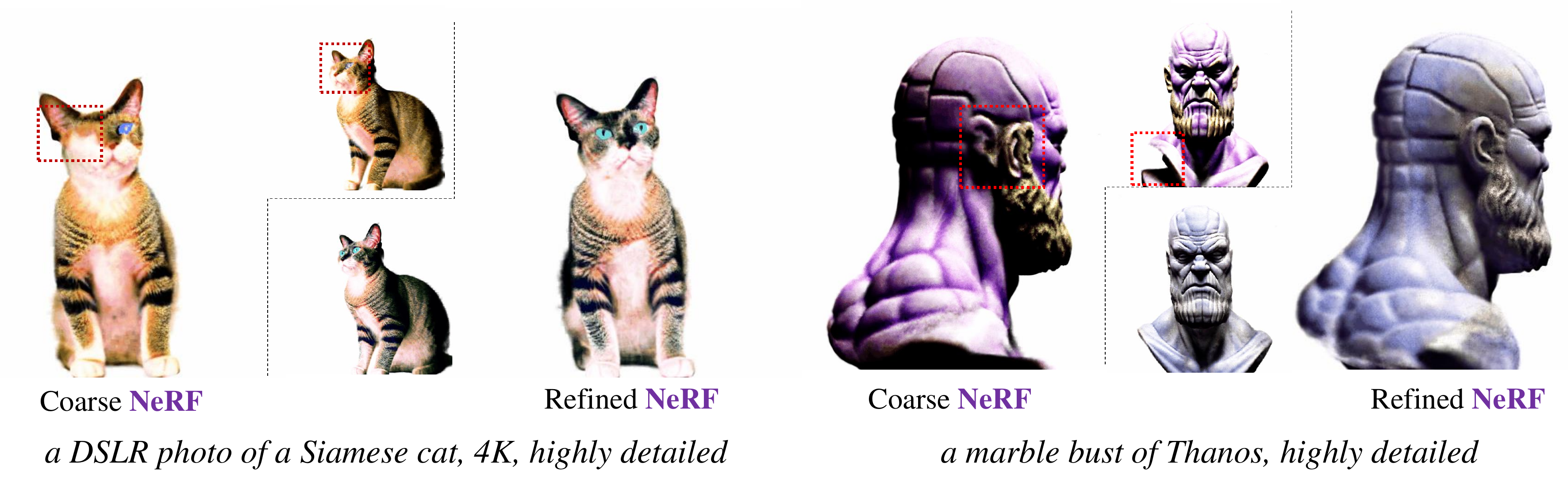}
    \caption{\textbf{IT3D can recover NeRF with mild geometry or Janus problem.} Current text-to-3D methods are potentially to generate failure cases like below examples, our method are capable of recovering failure cases with mild geometry or Janus problem.}
    \label{fig:f4}
\end{figure*}
\begin{figure*}[h!]
    \centering
    \includegraphics[width=\linewidth,trim={0cm 4.0cm 0.0cm 0.0cm}, clip]{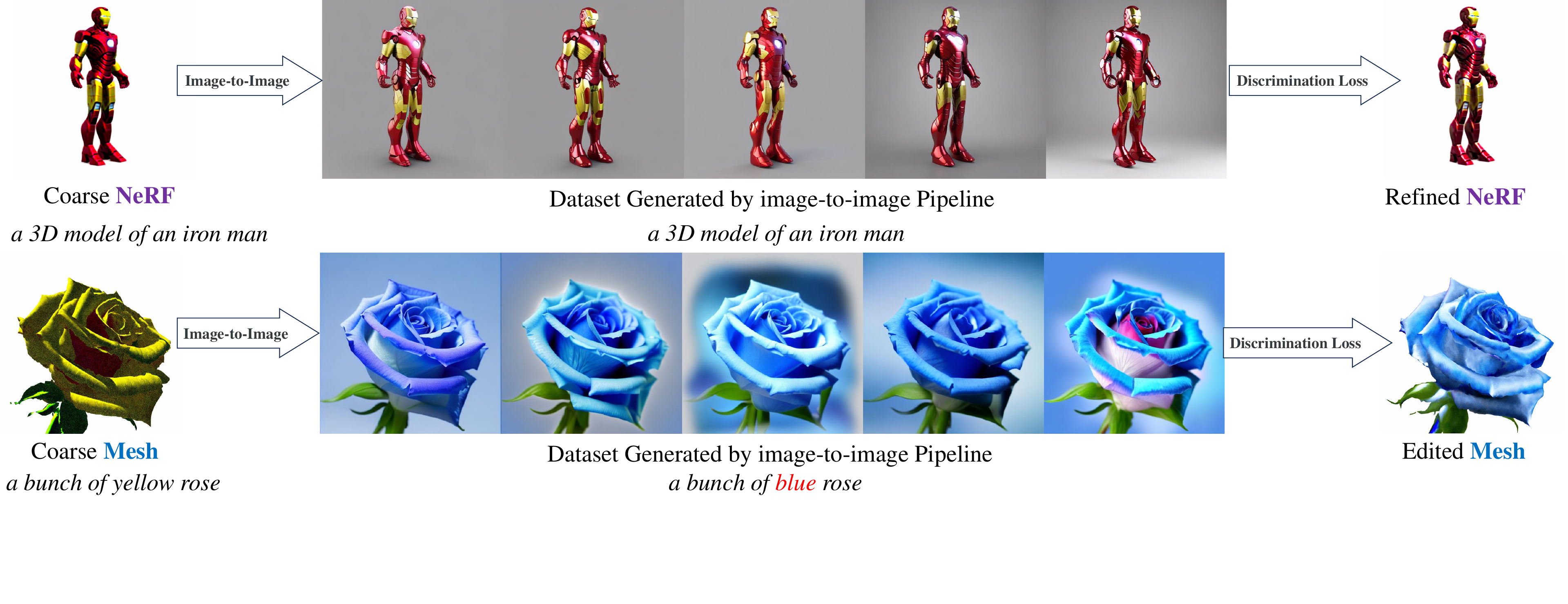}
    \caption{\textbf{Examples for generated dataset.} Our method can work on high-variance dataset and can tolerate failure cases of image-to-image pipeline(see texture of iron man dataset)}
    \label{fig:f5}
\end{figure*}
During the refinement phase, we fine-tune the coarse 3D model $\theta$ using both the discriminator loss and the diffusion prior. As the discriminator is randomly initialized, the discriminator loss introduces a considerable amount of noise along with meaningful content to the 3D model. However, due to the robust denoising capabilities inherent in large diffusion models, the diffusion prior effectively eliminates the added noise without erasing the valuable information acquired from the discriminator.

We found that this training strategy significantly reduce the training burden of the 3D GAN. Training a 3D GAN to achieve full convergence is highly time-consuming~\cite{chan2022efficient}. However, the proposed training strategy accomplishes full convergence within a mere 0.5 to 1.0 GPU hours. This abbreviated refinement period, however, is insufficient for the discriminator to thoroughly comprehend the nuances of fine textures and geometry, as seen in result (b) in Fig.~\ref{fig:f2}, which was trained for over 10.0 GPU hours. Nonetheless, the effectiveness of our training strategy is empirically validated.

We attribute this efficiency to the robust denoising capabilities and content dreaming capacity of the diffusion prior. Leveraging the diffusion prior empowers the generator to closely track the discriminator's updates at an accelerated pace. As a result, the discriminator can focus on learning higher-level distribution discrepancies without spending time on mastering minute details. Consequently, both the generator and discriminator update at significantly higher speeds compared to traditional vanilla 3D GANs.

\section{Experiments}

\subsection{Implementation details}

As our baseline method, we opt for the default configuration of text-to-3D in the Stable DreamFusion repository~\cite{stable-dreamfusion}. This baseline approach embraces classical SDS loss and the regularization losses outlined in DreamFusion~\cite{poole2022dreamfusion}. It is noteworthy that our IT3D method serves as a plug-and-play refinement technique and  can also be seamlessly applied to any text-to-3D method based on 2D diffusion priors.

For our baseline method, complex prompts such as avatars typically require approximately 20k steps to achieve complete convergence, while simpler prompts like flowers converge around 15k steps. To ensure a fair comparison, we train the baseline method for 25K steps (1.5 to 2.5 GPU hours) to guarantee the full convergence of the 3D model. As for our method, we resume training from 10k to 20k steps of the baseline method, varying based on the complexity of the prompt. Following the resumption and dataset generation steps, we proceed to initialize a discriminator using the same architecture as EG3D, commencing our Diffusion-GAN dual training strategy.

It is important to note that the entire refinement process is iterative in nature. After enhancing the 3D model's quality, the same process of dataset generation and refinement can be iteratively applied to the improved 3D model, resulting in further quality improvements. Additionally, the incorporation of a Dataset Update (DU) strategy~\cite{shao2023control4d,haque2023instruct} can complement our training approach. But in this experiment, we only generate one fixed dataset for each 3D model.

\subsection{3D Model Refinement and Editing}
In this section, we present our results for 3D model refinement and editing, as depicted in Fig.~\ref{fig:f1} and Fig.~\ref{fig:f3}.

We demonstrate that even with a low-quality coarse NeRF (Fig.~\ref{fig:f4}) and a generated dataset that includes failure cases (Fig.~\ref{fig:f5}), IT3D is capable of achieving a significant enhancement in quality.

In Fig.~\ref{fig:f4}, IT3D demonstrates success in rectifying slight Janus problems and effectively corrects erroneous geometry, such as character body poses.

\subsection{User Study}
We conduct a user study with the baseline method over 15 prompts. The renderings utilized for this study align with the settings employed in the earlier experiments. The study involved 45 participants, resulting in 675 pairwise comparisons. On average, our methods were preferred by 89.92\% in comparison to the baseline method.

\section{Conclusion}
In this paper, we propose a new plug-and-play refinement method for text-to-3D generation. Given a coarse 3D model, IT3D first generates posed image datasets by leveraging an image-to-image diffusion pipeline. It then refines the coarse 3D model with a novel Diffusion-GAN training strategy. To the best of our knowledge, our work is the first to combine GAN and diffusion prior to improve the text-to-3D task. 

\textbf{Limitations.} The performance of our proposed method is limited by the performance of image-to-image pipelines. Current image-to-image pipelines fail to generate high-quality results when the input prompt is too complicated.

{\small
\bibliographystyle{ieee_fullname}
\bibliography{egbib}
}
\clearpage
\appendix
\begin{center}
\textbf{\large Appendix}
\end{center}
\section{Introduction}
The content of our supplementary material is organized as follows:
\begin{itemize}
  \item Firstly, we provide more implementation details in  Section~\textbf{Implementation Details}.
  \item Secondly, in Section~\textbf{More Results}, we present additional experiment results with analysis.

\end{itemize}

\section{Implementation Details}
\subsection{Pipeline Details}
As mentioned in the main body of IT3D, we use the default configuration in the Stable DreamFusion repository~\cite{stable-dreamfusion} as our baseline method. Additionally, we adhere to the default loss weight settings provided by the same repository. After resuming from the checkpoint of the baseline method, the duration of the refining stage varies between 5k and 10k steps, corresponding to a time span of 0.5 to 1.0 GPU hours.

For both the baseline and IT3D methods, we perform rendering in the RGB color space with a rendering resolution of 512 to facilitate high-resolution synthesis. We sample 384 points along each rendering ray across all of our experiments. Due to memory limitations, the batch size is set to 1. Additionally, we employ Perp-Neg~\cite{armandpour2023reimagine} to mitigate the Janus problem.

For textured mesh generation, we adopt the default DMTET configuration~\cite{shen2021deep,chen2023fantasia3d} from the Stable DreamFusion repository. Both IT3D and the baseline approach initialize the DMTET with the coarse NeRF trained from the baseline method.

\subsection{GAN Details}
For the newly incorporated discriminator, we adopt a similar architecture, regularization function, and loss weight as the EG3D model~\cite{chan2022efficient}, with some distinctions.

Firstly, the vanilla structure of a 3D GAN involves a 3D generator that incorporates a super-resolution module, accompanied by a discriminator that accepts both coarse and fine image inputs~\cite{chan2022efficient}. Notably, due to the contextual disparities between text-to-3D and 3D GAN applications, we opt to omit the super-resolution component from the 3D GAN architecture. This choice stems from the consistent need to extract mesh or voxel representations from the 3D model within the context of text-to-3D.

Secondly, we gradually reduce the weight of the discrimination loss applied to our model, ultimately reaching zero weight towards the end of the training process. This weight decay strategy aims to allow the diffusion prior to complete the final refinement. As emphasized in the main paper, while the discrimination loss guides the update direction of the diffusion prior, the ultimate realistic texture and geometry are provided by the diffusion prior itself.

\begin{figure*}[!]
    \centering
    \includegraphics[width=\linewidth,trim={0cm 0.5cm 0.0cm 0.0cm}, clip]{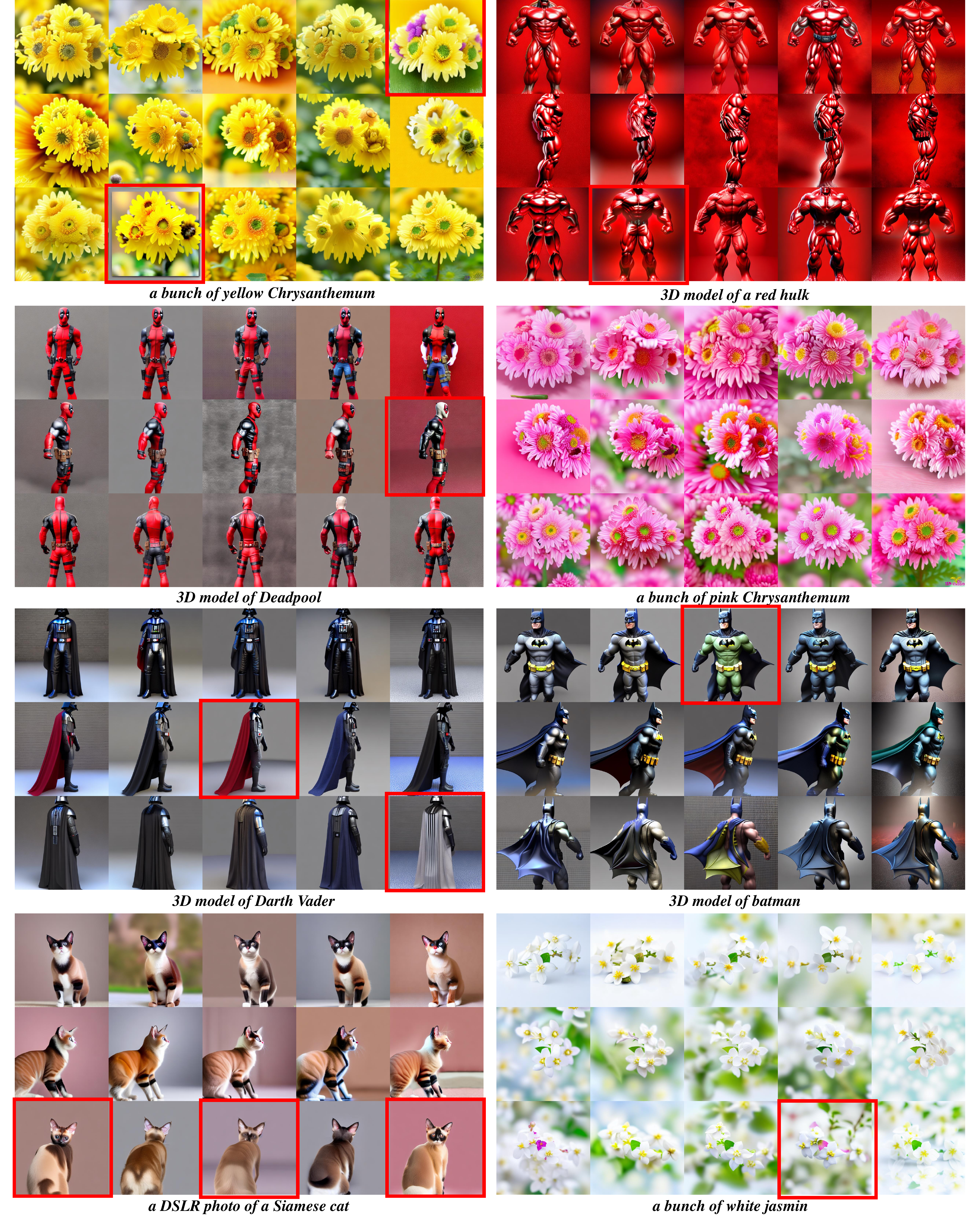}
    \caption{
    \textbf{Generated Dataset of IT3D.}
    We present three views for each prompt: front view, side view, and back view, arranged from top to bottom.  As highlighted by the red squares, the generated datasets include  failure cases that are that are misaligned with image and prompt conditions. This demonstrates that our approach is capable of accommodating such failure generation cases.}
    \label{fig:sup2}
\end{figure*}

\section{More Results}
As depicted in Figure~\ref{fig:sup2}, IT3D doesn't strictly require a high-quality dataset. For example, front view samples are incorrectly generated when conditioned on back view renderings (hulk and Siamese cat). Additionally, there are cases of prompt misalignment  (batman and Darth Vader). This illustrates the adaptability of our approach to cope with such instances of generation failure.
\end{document}